\pdfoutput=1

\documentclass[11pt]{article}

\usepackage{EMNLP2023}
\usepackage{times}
\usepackage{latexsym}

\usepackage{graphicx}
\usepackage[T1]{fontenc}

\usepackage[utf8]{inputenc}

\usepackage{microtype}

\usepackage{inconsolata}

\usepackage{booktabs}
\usepackage{times}
\usepackage{latexsym}
\usepackage{booktabs}
\usepackage{hyperref}

\usepackage{graphicx}
\usepackage[T1]{fontenc}

\usepackage[utf8]{inputenc}

\usepackage{microtype}

\usepackage{inconsolata}

\newcommand{\bv}{\mathbf{V}}
\newcommand{\bx}{\mathbf{x}}

\newcommand{\bl}{\mathbf{L}}

\title{XrayGPT: Chest Radiographs Summarization using Large Medical Vision-Language Models}

\author{%
\\Omkar Thawakar*$^{1}$ \quad 
  Abdelrahman Shaker*$^{1}$ \quad 
  Sahal Shaji Mullappilly*$^{1}$ \quad 
  Hisham Cholakkal$^{1}$ \\
  Rao Muhammad Anwer$^{1,2}$ \quad
  Salman Khan$^{1}$ \quad
  Jorma Laaksonen$^{2}$ \quad
  Fahad Shahbaz Khan$^{1}$
  \vspace{0.5em} \\
  $^{1}$Mohamed bin Zayed University of AI \quad 
  $^{2}$Aalto University  \quad
  }

\begin{document}
\maketitle
\begin{NoHyper}
\def\thefootnote{*}\footnotetext{First Author Equal Contribution}
\end{NoHyper}
\begin{abstract}
The latest breakthroughs in large vision-language models, such as Bard and GPT-4, have showcased extraordinary abilities in performing a wide range of tasks.
Such models are trained on massive datasets comprising billions of public image-text pairs with diverse tasks. However, their performance on task-specific domains, such as radiology, is still under-investigated and potentially limited due to a lack of sophistication in understanding biomedical images. 
On the other hand, conversational medical models have exhibited remarkable success but have mainly focused on text-based analysis.
In this paper, we introduce XrayGPT, a novel conversational medical vision-language model that can analyze and answer open-ended questions about chest radiographs. Specifically, we align both medical visual encoder (MedClip) with a fine-tuned large language model (Vicuna), using a simple linear transformation. 
This alignment enables our model to possess exceptional visual conversation abilities, grounded in a deep understanding of radiographs and medical domain knowledge. 
To enhance the performance of LLMs in the medical context, we generate ~217k interactive and high-quality summaries from free-text radiology reports. These summaries serve to enhance the performance of LLMs through the fine-tuning process. Our approach opens up new avenues the research for advancing the automated analysis of chest radiographs. Our open-source demos, models, and instruction sets are available at: \url{https://github.com/mbzuai-oryx/XrayGPT}
\end{abstract}

\section{Introduction}
The Large-scale Vision-Language models have emerged as a transformative area of research at the intersection of computer vision and natural language processing, enabling machines to understand and generate information from both visual and textual modalities. These models represent a significant advancement in the field, bridging the gap between visual perception and language comprehension, and have demonstrated remarkable capabilities across various tasks, including but not limited to image captioning~\cite{hossain2019comprehensive}, visual question answering~\cite{lu2023multiscale}, and visual commonsense reasoning~\cite{zellers2019recognition}. Training these models requires vast amounts of image and text data, enabling them to learn rich representations that capture the intricacies of both modalities. Additionally, fine-tuning can be employed using task-specific data to better align the models with specific end tasks and user preferences. Recently, Bard and GPT-4 have demonstrated impressive capabilities in various tasks, raising excitement within the research community and industry. However, it is important to note that the models of Bard and GPT-4 are not currently available as open-source, limiting access to their underlying architecture and implementation details. 

Recently, Mini-GPT~\cite{minigpt} demonstrates a range of impressive capabilities by aligning both vision and language models. It excels at generating contextual descriptions based on the given image. However, it is not as effective in medical scenarios due to the significant differences between medical image-text pairs and general web content. Adopting vision-text pre-training in the medical domain is a challenging task because of two factors: (1) Lack of data, Mini-GPT has trained the projection layer on a dataset of 5M image-text pairs, while the total number of publicly available medical images and reports is orders of magnitude below. (2) Different modalities and domains, while Mini-GPT may involve distinguishing between broad categories like "Person" and "Car" the distinctions within medical domains are much more subtle and fine-grained. For instance, differentiating between terms like "Pneumonia" and "Pleural Effusion" requires more precision by capturing and aligning crucial medical domain knowledge.

Chest radiographs are vital for clinical decision-making as they offer essential diagnostic and prognostic insights about the health of the patients. Text summarization tasks can partially address this challenge by providing meaningful information and summaries based on the given radiology reports. In our approach, we go beyond traditional summarization techniques by providing concise summaries that highlight the key findings and the overall impression based on the given X-ray. Additionally, our model allows for interactive engagement, enabling users to ask follow-up questions based on the provided answers. In this study, we stimulate the research around the automated analysis of chest radiographs based on X-ray images. Also, we argue that based on the visual and large language models, the majority of knowledge acquired during the pertaining stage of these models requires a domain-specific high-quality instruction set derived from task-specific data to achieve promising results. The main contributions of our work are:-
\begin{itemize}\setlength{\itemsep}{0em}
\item 
The LLM (Vicuna) is fine-tuned on medical data (100k real conversations between patients and doctors) and ~20k radiology conversations to acquire domain-specific and relevant features. 

\item 
We generate interactive and clean summaries (~217k) from free-text radiology reports of two datasets: MIMIC-CXR~\cite{MIMIC_CXR} and OpenI~\cite{OpenI_dataset}. These summaries serve to enhance the performance of LLMs by fine-tuning the linear transformation layer on high-quality data.

\item 
We align the frozen specialized medical visual encoder (MedClip) with a fine-tuned LLM (Vicuna), using a simple linear transformation to understand medical meanings and acquire remarkable visual conversation abilities.
\item 
To advance the research in biomedical multimodal learning, we open-source our assets to the community: The codebase, the fine-tuned models, the high-quality instruction-set, and the raining recipe for data generation and model training are publically released.
\end{itemize}

\section{Related Work}

\noindent\textbf{Medical Chatbot} Medical chatbots have emerged as valuable tools in healthcare, providing personalized support, information, and assistance to patients and healthcare professionals. The recently introduced Chatdoctor~\cite{yunxiang2023chatdoctor} is a next-generation AI doctor model that is based on the LLaMA~\cite{touvron2023llama} model. The goal of this project is to provide patients with an intelligent and reliable healthcare companion that can answer their medical queries and provide them with personalized medical advice. After success of ChatGPT~\cite{chatgpt}, GPT-4~\cite{openai2023gpt4} and other open source LLM's \cite{touvron2023llama,vicuna2023,alpaca}, many medical chatbots were introduced recently such as Med-Alpaca~\cite{han2023medalpaca}, PMC-LLaMA~\cite{wu2023pmc}, and DoctorGLM~\cite{xiong2023doctorglm}. These models utilize open source LLM's and finetuned on specific medical instructions. These studies highlight the potential of medical chatbots to improve patient engagement and health outcomes through interactive and personalized conversations. Overall, medical chatbots offer promising opportunities using only textual modality for enhancing AI healthcare, with ongoing research focused on refining their capabilities with ethical considerations.

\noindent\textbf{Large Language Vision Models} A significant area of research in natural language processing (NLP) and computer vision is the exploration of Large Language-Vision Model (LLVM) learning techniques. This LLVM aims to bridge the gap between visual and textual information, enabling machines to understand and generate content that combines both modalities. Recent studies have demonstrated the potential of LLVM models in various tasks, such as image captioning \cite{minigpt}, visual question answering~\cite{bazi2023vision,liu2023q2atransformer,Maaz2023VideoChatGPT}, and image generation~\cite{zhang2023adding}.

\section{Method}
XrayGPT is an innovative conversational medical vision-language model specifically developed for analyzing chest radiographs. The core concept revolves around aligning medical visual and textual representations to enable the generation of meaningful conversations about these radiographs using our generated high-quality data. Our approach draws inspiration from the design of vision-language models in general, but with a specific focus on the medical domain. Due to the limited availability of medical image-summary pairs, we adopt a similar methodology by building upon a pre-trained medical vision encoder (VLM) and medical large language model (LLM), as our foundation. The fine-tuning process involves aligning both modalities using high-quality image-summary pairs through a simple transformation layer. This alignment enables XrayGPT to possess the capability of generating insightful conversations about chest radiographs, providing valuable insights for medical professionals. By leveraging pre-existing resources and fine-tuning specific components, we optimize the model's performance while minimizing the need for extensive training on scarce data.

\subsection{Model Architecture}
\begin{figure*}[t]
  \centering
    \includegraphics[width=0.96\linewidth]{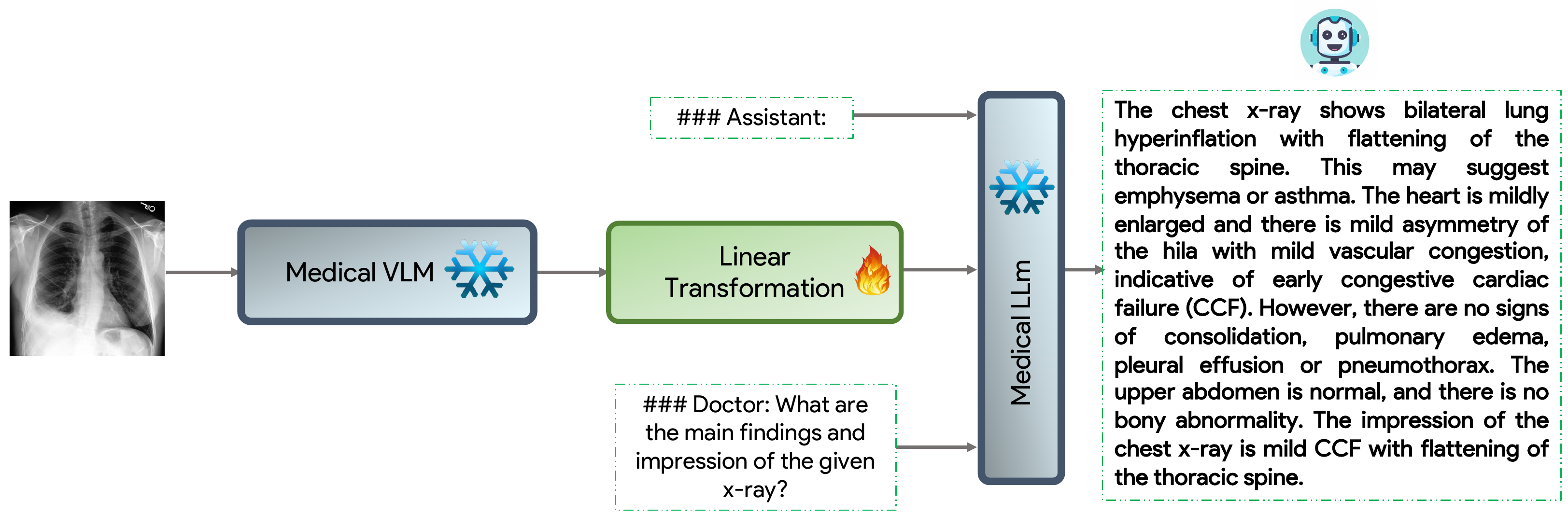}
    \caption{Overview of our XrayGPT framework. The input X-ray is passed sequentially to three components. (1) Frozen medical visual encoder to extract relevant features pertaining to the chest diagnosis. (2) Leanable linear transformation layer to align the medical visual features with the Medical LLM together to learn extensive medical visual-text alignment. (3) Frozen Medical LLM to generate a detailed summary of the X-ray based on the encoded features and the given prompt.}
    \label{fig:Architecture}
    \vspace{-0.5cm}
\end{figure*}

We show in Fig. \ref{fig:Architecture} an overview of our XrayGPT. Given the X-ray, we align both visual features and textual information from a pre-trained medical vision encoder (VLM), and medical large language model (LLM). Specifically, we utilize MedClip~\cite{medclip} as a visual encoder and our large language model (LLM) is built upon the recent Vicuna~\cite{vicuna2023}.

Given X-ray $\bx \in {R}^{H \times W  \times C}$, the visual encoder is used to encode the image into embeddings using a vision encoder $E_{img}$. Then, the raw embeddings are mapped to an output dimension of 512 using a linear projection head.
\begin{equation}
\bv_p = f_v(E_{img}(\bx))
\end{equation}
where $E_{img}$ is the vision encoder, $f_v$ is the projection head.

To bridge the gap between image-level features and the language decoder's embedding space, we employ a trainable linear transformation layer, denoted as $t$. This layer projects the image-level features, represented by $\bv_p$, into corresponding language embedding tokens, denoted as $\bl_v$:

\begin{equation}
\bl_v = t(v_p),
\end{equation}

We have two text queries in our overall architecture. The first query, denoted as \#\#\#Assistant, serves the purpose of determining the system role, which in our case is defined as "\textit{You are a helpful healthcare virtual assistant}." The second text query, \#\#\#Doctor, corresponds to the prompt itself. To ensure consistency, both queries undergo tokenization, resulting in dimensions represented by $\bl_t$. Finally, $\bl_v$ is concatenated with $\bl_t$ and fed into the medical LLM, Fine-tuned Vicuna, which generates the summary of the chest x-ray.

Our XrayGPT follows a two-stage training approach. In the first stage, we train the model using interactive summaries from MIMIC-CXR~\cite{MIMIC_CXR} reports. While in the second stage, we use the high-quality curated interactive summaries of OpenI~\cite{OpenI_dataset} reports. MIMIC-CXR report findings contain information about patient history, which adds noise to the data. To mitigate this effect, we use small yet effective Interactive OpenI report summaries in the second stage to make our model robust to noise.

\subsection{Image-text alignment}
To align the generated high-quality summaries with the given x-ray, we use similar conversational format of the Vicuna \cite{vicuna2023} language model as follows:

\textit{\#\#\#Doctor: $ X_{R}X_{Q}$ \#\#\#Assistant: $X_{S}$}

where $X_{R}$ is the visual representation produced by the linear transformation layer for image $X$, $X_{Q}$ is a sampled question (e.g. What are the main findings and impression of the given X-ray?), and $X_{S}$ is the associated summary for image $X$. In this way, we curate image-text pairs with detailed and informative interactive summaries.

\section{Curating high-quality data}
\label{data:prepare}

\noindent\textbf{Datasets:} The MIMIC-CXR consists of a collection of chest radiographs associated with free-text radiology reports. It consists of 377,110 images and 227,827 associated reports, which are used for both training and testing purposes. The dataset is de-identified by removing the health information to satisfy health insurance and privacy requirements. The OpenI dataset is a collection of chest X-ray images from the Indiana University hospital network, composing 6,459 images and 3,955 reports.

\noindent\textbf{High Quality and Interactive Summaries:} To generate concise and coherent medical summaries from the unstructured reports, we perform the following pre-processing steps for both datasets: (1) Removal of incomplete reports lacking finding or impression sections. (2) Elimination of reports that have finding sections containing less than 10 words. (3) Exclusion of reports with impression sections containing less than 2 words.

In addition, utilizing the power of gpt-3.5-turbo model, we further implement the following pre-processing techniques to ensure high-quality summaries per image: (1) Elimination of sentences containing comparisons to the patient's prior medical history. (2) Removal of de-defined symbols "\_\_", while preserving the original meaning. (3) As our training relies on image-text pairs, we excluded the provided view from the summary. (4) We combine the clean findings and impressions to generate an interactive and high-quality summary.

Following these steps, we obtained a set of filtered training reports consisting of 114,690 reports associated with 241k training images based on Mimic-CXR dataset. Also, we obtained 3,403 high-quality summaries that used for training based on the OpenI dataset.

Here is an example before and after the proposed pre-processing. \textbf{Input findings}:
\textit{PA and lateral views of the chest were provided demonstrating no focal consolidation, effusion, or pneumothorax. Cardiomediastinal silhouette appears normal and stable. There is a compression deformity involving a mid thoracic vertebral body, which appears new from the prior chest radiograph of \_\_\_. No free air below the right hemidiaphragm. There are tiny surgical clips in the left base of neck, likely indicating prior thyroid surgery}. \textbf{Input Impression}: \textit{No acute intrathoracic process. Interval development of a mid thoracic spine compression fracture.} \textbf{High-quality and interactive summary}: \textit{The chest x-ray findings reveal no evidence of focal consolidation, effusion, or pneumothorax. The cardiomediastinal silhouette appears stable and normal. There is a newly developed mid thoracic spine compression fracture but no free air below the right hemidiaphragm. The presence of surgical clips in the left base of the neck suggests prior thyroid surgery. The impression suggests that there is no acute intrathoracic condition detected in the x-ray aside from the new development of mid thoracic spine compression fracture.}

\section{Experiments}

\subsection{Implementation Details}

\noindent\textbf{Stage-1 Training:} In stage-1 training, the model is designed to gain understanding of how Xray image features and corresponding reports are interconnected by analysing a large set of image-text pairs. The result obtained from the injected projection layer is considered as a gentle cue for our medically tuned LLM model, guiding it to produce the appropriate report based on the finding and impression that match the given x-ray images. We use high quality interactive report summary as described in sec.~\ref{data:prepare} of MIMIC-CXR~\cite{MIMIC_CXR} train set with 213,514 image text pairs for training. During training the model trained for 320k total training steps with a total batch size of 128 using 4 AMD MI250X (128GB) GPUS.    

\noindent\textbf{Stage-2 Training:} In stage-2 training, the pretrained stage-1 model is enforced to gain radiology specific summary of how xray image features by examining set of highly curated image-text pairs from OpenI dataset. As a result, our medically tuned LLM can produce more natural and high quality radiology specific responses of given chest Xray images. We use high quality interactive report summary as described in sec.~\ref{data:prepare} from OpenI~\cite{OpenI_dataset} set with ~3k image text pairs for training. During training the model trained for 5k total training steps with a total batch size of 32 using single AMD MI250X (128GB) GPU.

In both stage-1 and stage-2 of training, we utilize predetermined prompts in the given format:

\#\#\#Doctor: <Img><ImageFeature></Img> <Instruction> \#\#\#Assistant:

Here, <Instruction> refers to a randomly selected instruction from our pre-established set of instructions, which includes different forms of instructions like "Describe the given chest x-ray image in detail." or "Are there any potential complications or risks associated with the observed abnormalities in this chest x-ray image? or the x-ray is normal." or "Is the overall impression provided by this chest x-ray image normal or abnormal? Answer based on the observed findings.". Similar to the baseline~\cite{minigpt}, we do not compute the regression loss for this particular text-image prompt.

\subsection{Evaluation Metrics}
\label{eval_metrics}

We used the Rogue Score as an evaluation metric to compare the contribution of our components over the baseline~\cite{minigpt}. Rogue score has been commonly used ~\cite{cheng2016neural,nallapati2017summarunner,zhang2020pegasus} to assess the quality of generated text, particularly in the field of natural language processing and text generation. It measures the overlap between the generated text and a set of reference texts, typically generated by human experts. The Rogue Score calculates precision, recall, and F1-score, taking into account the presence and ordering of n-grams (contiguous sequences of words). A higher Rogue Score indicates a better alignment between the generated text and the reference texts, indicating a higher level of quality and coherence. The Rogue Score serves as a valuable quantitative measure to objectively compare and assess the performance of different text generation models and techniques.

GPT-based evaluation of LMM (Language Model Mediated) generated text refers to the use of GPT (Generative Pre-trained Transformer) models to assess the quality and coherence of text generated by LMM approaches. LMM combines the power of pre-trained language models, such as GPT, with explicit control mechanisms to guide text generation. GPT-based evaluation involves using a fine-tuned GPT model to generate a set of reference texts, which can then be compared with the LMM-generated texts. Metrics such as perplexity, BLEU score, or the recently proposed Self-BLEU score can be used to quantitatively evaluate the similarity between the reference and generated texts. 

\subsection{Quantitative Measures}

\begin{table}
\centering
  \scalebox{0.98}{
  \begin{tabular}{l|c|c|c}
    \toprule
    \textbf{Method} & \textbf{R-1} & \textbf{R-2} & \textbf{R-L} \\
    \midrule
    Baseline & 0.1313 & 0.0221 & 0.0879 \\
    + MedCLIP & 0.1517 & 0.0308 & 0.0973 \\
    + MedVicuna & 0.2099 & 0.0551 & 0.1284 \\
    + RadVicuna & \textbf{0.3213} & \textbf{0.0912} & \textbf{0.1997} \\
    \bottomrule
  \end{tabular}
 }
 \caption{Comparison of our XrayGPT components with Baseline~\cite{minigpt} using Rogue score on MIMIC-CXR~\cite{MIMIC_CXR} Test set. All results are reported by progressively adding our components in the baseline Minigpt-4~\cite{minigpt}. Our approach outperforms the recent Minigpt-4 with an absolute gain of 19\% in terms of R-1 score. Best results are in bold.
 }
  \label{table:rogue_scores}
\end{table}

In this section, we highlight a key contribution of our XrayGPT compared to our baseline~\cite{minigpt}. We conduct quantitative evaluation using advanced metrics such as Rogue score and GPT-based evaluation as described in sec.~\ref{eval_metrics}. Tab.~\ref{table:rogue_scores} shows comparison of our key components when progressively integrated into our baseline~\cite{minigpt} frame. From Tab.~\ref{table:rogue_scores} our XrayGPT (row 4) has a significant improvement of 19\% over the state-of-the-art baseline~\cite{minigpt} on the MIMIC-CXR test set. Also, we did LLM's based evaluation by asking ChatGPT model to choose "which response is closer to reference between baseline vs XrayGPT" where our model scored 82\% compared to baseline 6\% showing the superiority of our XrayGPT for radiology-specific summary. 

\subsection{Qualitative Examples}
We performed an extensive evaluation of our xrayGPT on a various Chest Xray samples where our model not only provides detailed findings and impression for given chest xray but also shows the conversation capability like radiologist. Please refer Figure 2 to 5 for more details.

\begin{figure*}
  \centering
    \includegraphics[width=0.96\linewidth]{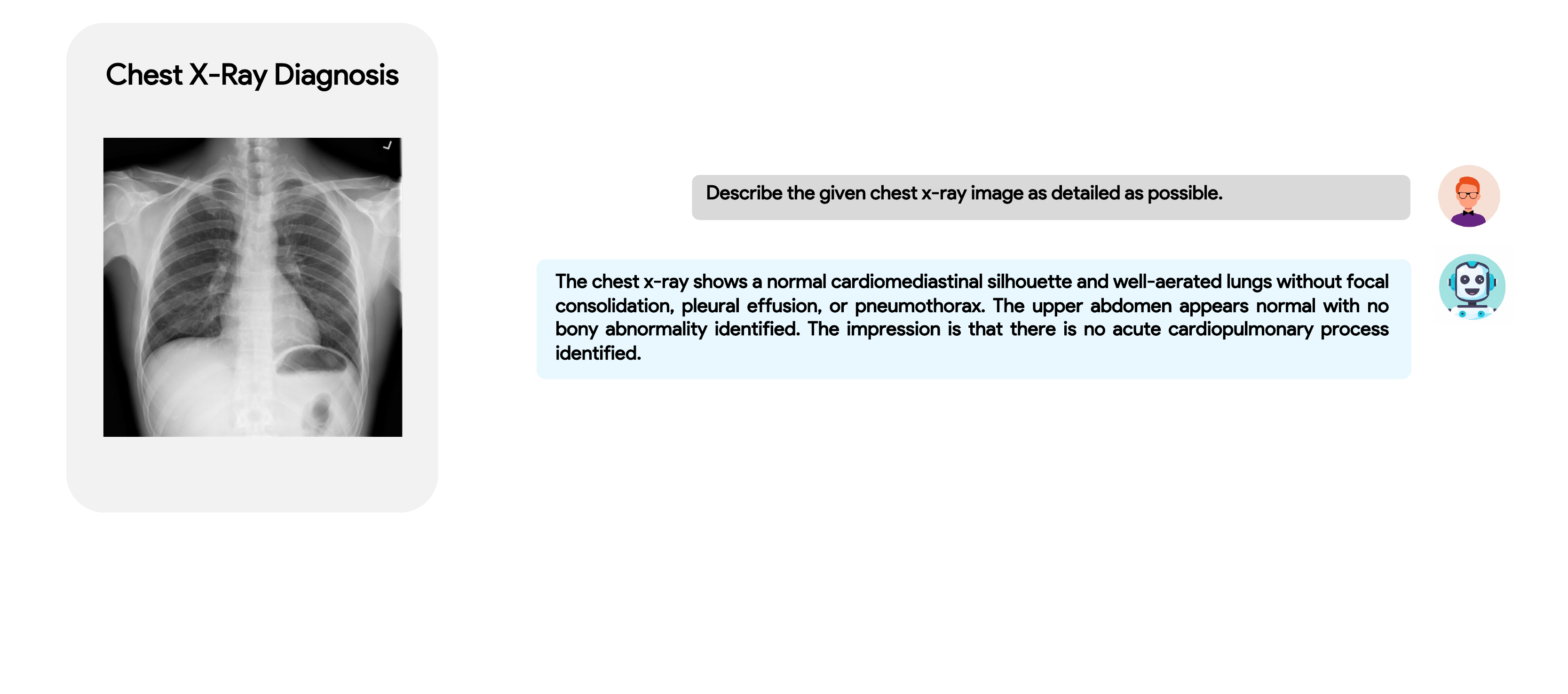}
    \caption{\textbf{XrayGPT: Delivering Comprehensive and Succinct Findings and Impressions for Every X-ray Image.} Our XrayGPT provides detailed radiology specific summary as well as concise findings and impression for given xray image.}
    \label{fig:example1}
    \vspace{-0.5cm}
\end{figure*}

\begin{figure*}
  \centering
    \includegraphics[width=0.96\linewidth]{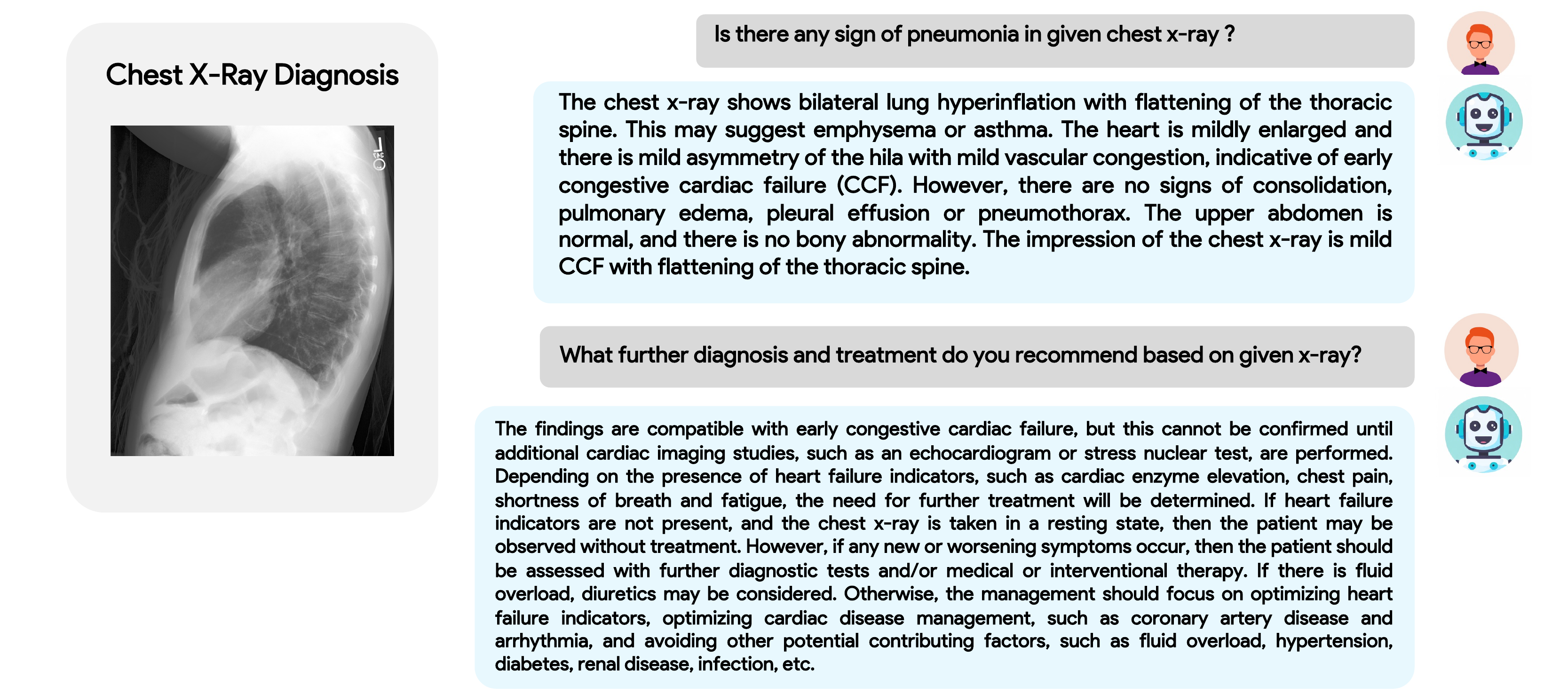}
    \caption{\textbf{XrayGPT: Empowering Users with Concise Insights - Explore the Comprehensive Analysis Derived from the xray Image.} Our XrayGPT provides a brief explanation for users quaries based on the overall findings from the given xray image.}
    \label{fig:example2}
    \vspace{-0.5cm}
\end{figure*}

\begin{figure*}
  \centering
    \includegraphics[width=0.96\linewidth]{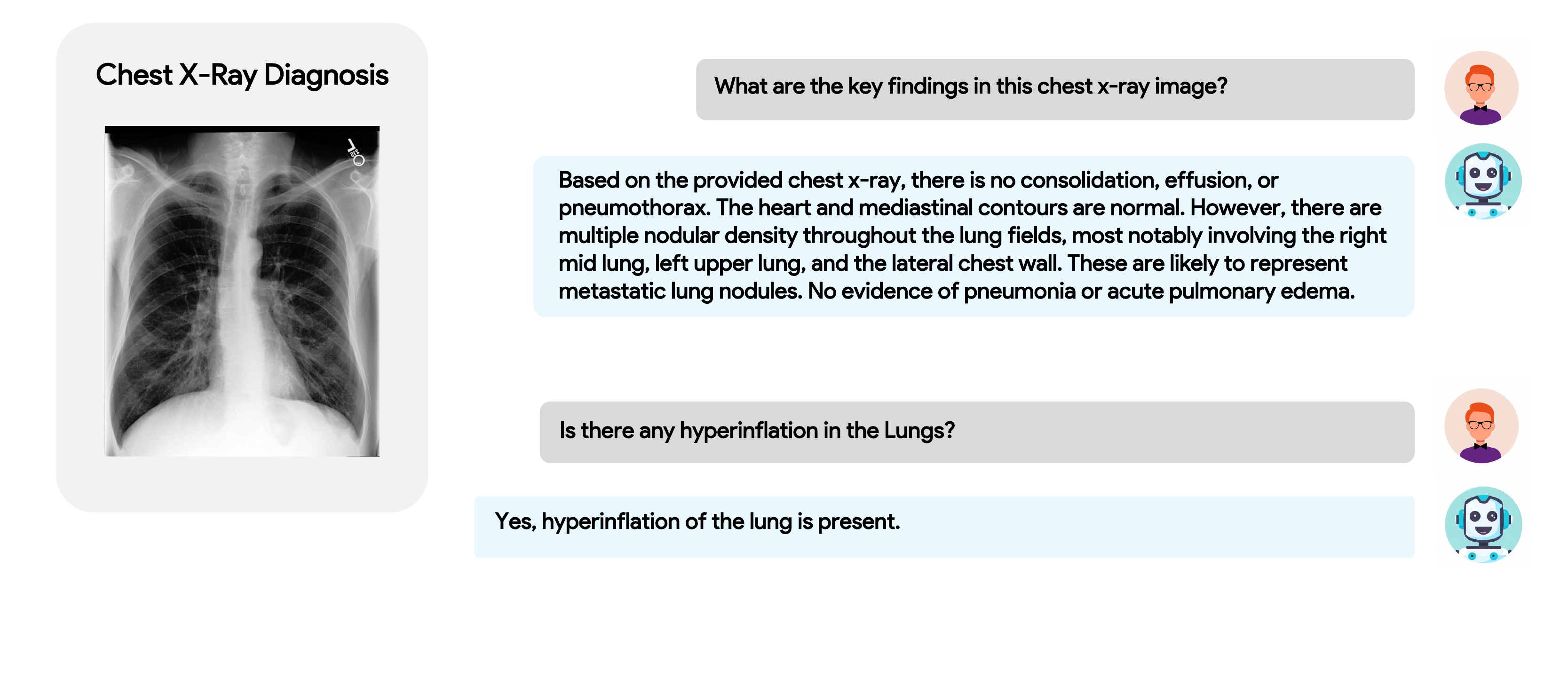}
    \caption{\textbf{XrayGPT: Unlocking the Power of Precision, Providing Specific Key Details in Response to User Queries.} Our XrayGPT is able to answer specific key detail in response to user queries.}
    \label{fig:example3}
    \vspace{-0.5cm}
\end{figure*}

\begin{figure*}
  \centering
    \includegraphics[width=0.96\linewidth]{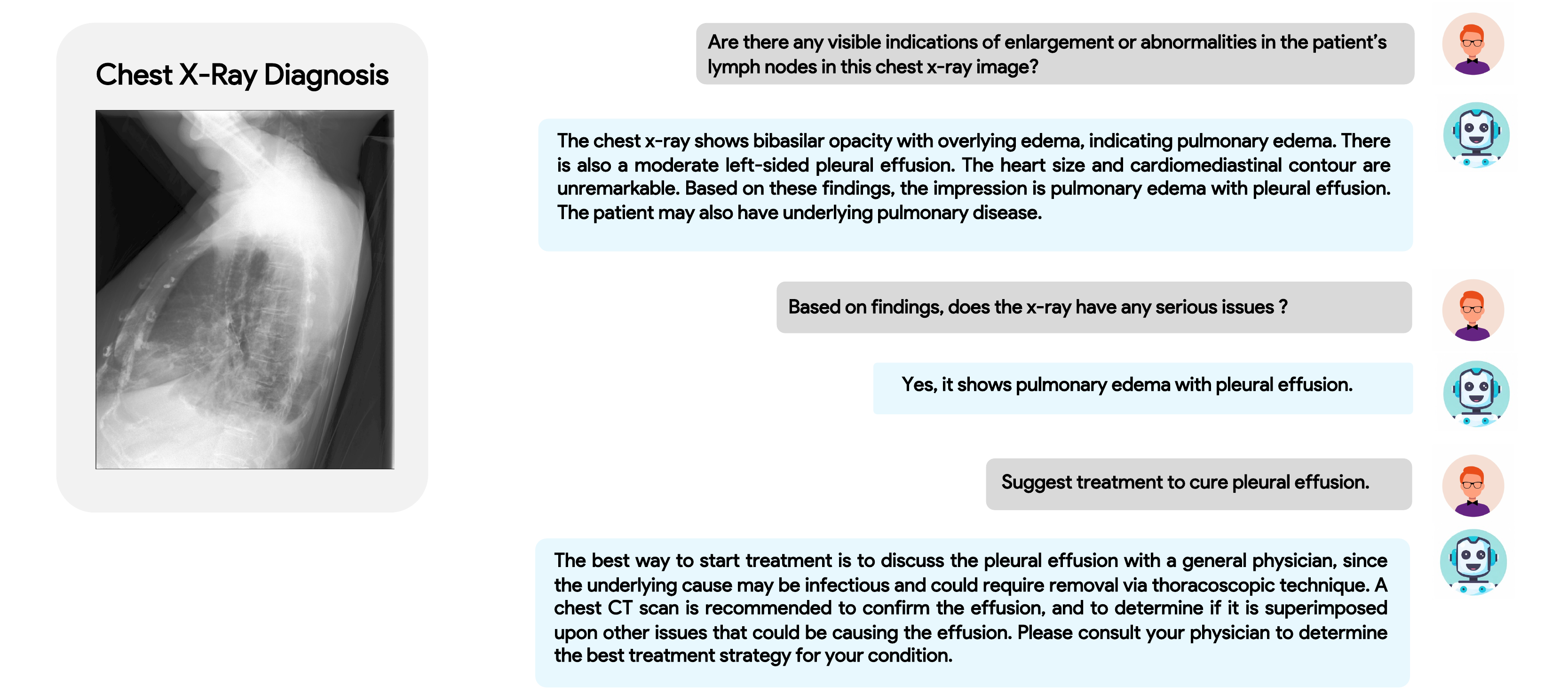}
    \caption{\textbf{XrayGPT: The Conversational AI Revolutionizing Radiological Interactions.} Our XrayGPT has radiological conversational capabilities.}
    \label{fig:example4}
    \vspace{-0.5cm}
\end{figure*}

\begin{figure*}
  \centering
    \includegraphics[width=0.96\linewidth]{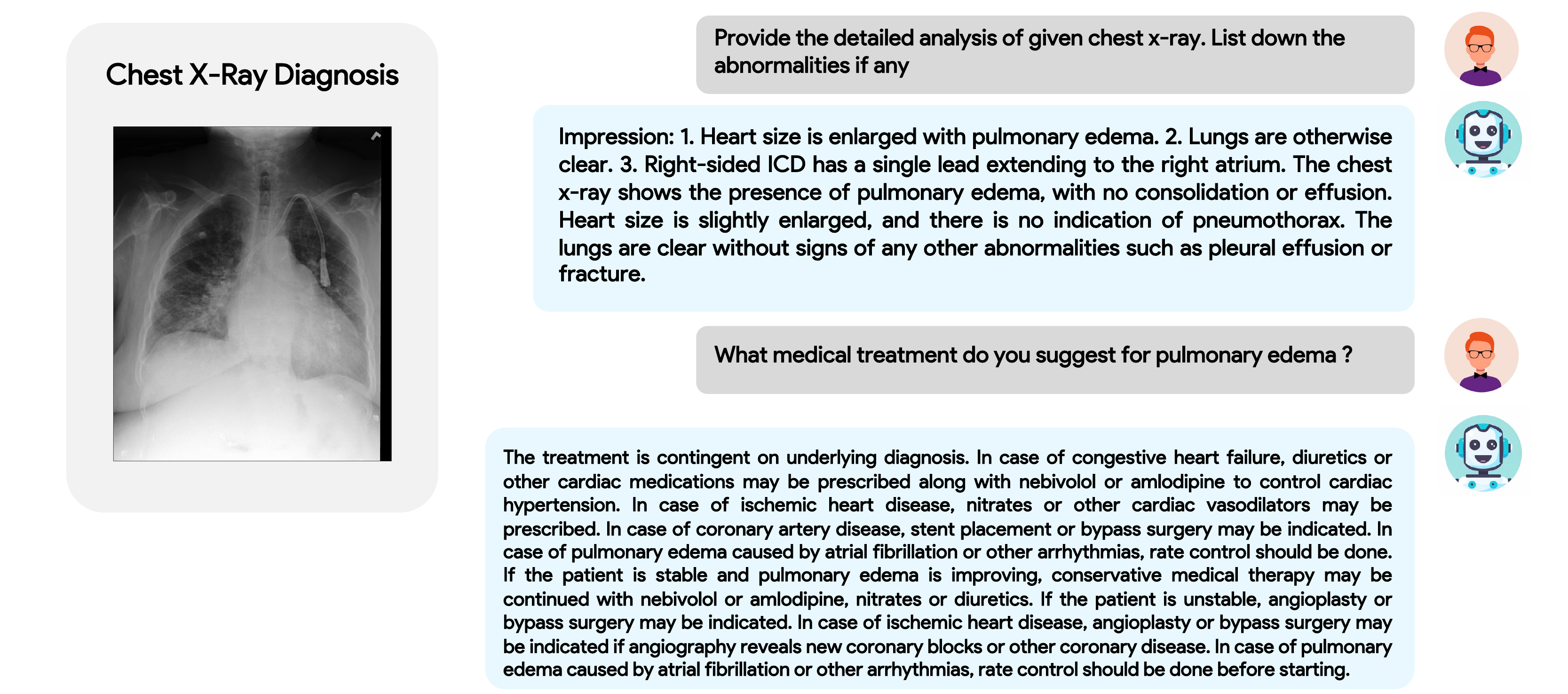}
    \caption{\textbf{XrayGPT: Medical Treatment Recommendation.} Our XrayGPT has the capability to suggest treatment based on the diagnosis.}
    \label{fig:example5}
    \vspace{-0.5cm}
\end{figure*}

\section{Conclusion}

To conclude, we present XrayGPT, a novel conversational medical vision-language model that combines both modalities to analyze and answer questions about chest radiographs. By aligning these models and leveraging our proposed interactive summaries from free-text radiology reports, XrayGPT demonstrates exceptional visual conversation abilities grounded in a deep understanding of chest radiographs in terms of findings and impressions. We open-source demos, models, and instruction sets. This work opens up new possibilities for advancing the automated analysis of chest radiographs and enhances the performance of large language models in the medical context.

\bibliography{emnlp2023}
\bibliographystyle{acl_natbib}

\appendix


\end{document}